%% file: paper.tex
\documentclass[runningheads]{llncs}
\usepackage[T1]{fontenc}
\usepackage{graphicx}
\usepackage{tablefootnote} 
\usepackage{multirow}
\usepackage{lscape}
\usepackage{hyperref}

\usepackage{color}

\AtBeginEnvironment{thebibliography}{\NoHyper}
\AtEndEnvironment{thebibliography}{\endNoHyper}

\begin{document}
%
\title{METATR: A Multilingual, Evolving Benchmark for Automatic Text Recognition}
\titlerunning{A Multilingual Benchmark for ATR}
%
\author{Mélodie Boillet\inst{1}\orcidID{0000-0002-0618-7852} \and
Solène Tarride\inst{1}\orcidID{0000-0001-6174-9865} \and
 Christopher Kermorvant\inst{1}\orcidID{0000-0002-7508-4080}}
\authorrunning{M. Boillet et al.}
%
\institute{TEKLIA, Paris, France\\
\email{\{mboillet, starride, kermorvant\}@teklia.com}\\
\url{https://www.teklia.com/}}
\maketitle              
\begin{abstract}

Benchmarks that reflect the diversity and complexity of real-world documents are essential for accurately evaluating Automatic Text Recognition (ATR) systems, especially Vision-Large Language Models (vLLMs). Although recent models demonstrate impressive performance, they are often evaluated on datasets containing modern, printed texts mostly written in English, which limits their relevance to many practical applications. Therefore, selecting a model for a specific use case requires evaluating it on data that matches the target documents. This highlights the importance of representative benchmarks for real-world applications. In this paper, we introduce METATR (v1.0), a multilingual, evolving benchmark designed to evaluate ATR models across a wide range of documents, facilitating meaningful model comparison and selection.

The benchmark was designed to maximize diversity by including documents from various public collections. These documents cover 29 languages and include texts with multiple scripts and layouts. Beyond the dataset itself, METATR defines a standardized prompting and normalization methodology and establishes a dynamic evaluation framework. This approach is intended to produce reproducible results while remaining extensible over time.

We evaluated a wide range of state-of-the-art systems, including open-source models, such as Churro and PeroOCR, as well as closed-source models, such as Gemini 3 and Claude. Results are reported across various dimensions, including performance at the dataset and language levels, robustness to handwritten documents, and computational efficiency. Our findings show that, although proprietary models achieve the most consistent performance, substantial variability persists across scripts and layouts. Overall, METATR provides a multidimensional, practitioner-oriented framework for assessing multilingual ATR in real-world conditions and tracking progress as the field evolves.


\end{abstract}

\section{Introduction}
\label{sec:introduction}

Automatic text recognition (ATR), which encompasses optical character recognition (OCR) and handwritten text recognition (HTR), has rapidly progressed with the rise of deep learning and large multimodal models. On standard benchmarks with clean, high-resolution images of printed or neatly handwritten text, the accuracy rate now exceeds 99\%. Recent frontier models have led to the claim that text recognition has been solved. These results suggest that general-purpose multimodal systems could replace specialized OCR and HTR pipelines.

However, such claims are largely based on a limited set of evaluation conditions. Most benchmarks focus on modern documents with simple layouts and short texts in a limited set of high-resource languages, mainly English and Chinese. In contrast, real-world collections from archives, libraries and cultural heritage institutions contain degraded documents, historical scripts, multilingual content, non-Latin alphabets, complex layouts, and long pages that require a coherent reading order. Under these conditions, performance often drops significantly, and issues such as hallucinations and reading-order errors become critical. As a result, high scores on standardized benchmarks do not necessarily translate into robust performance on heterogeneous historical documents.

Current historical document benchmarks are often shaped by the objectives of their creators. Those designed by end users, such as historians or archivists, typically reflect the needs of a specific project and therefore concentrate on a narrow range of documents, languages, or scripts. While these datasets are valuable, they offer limited diversity and generalizability. Conversely, benchmarks introduced by model developers may highlight the strengths of particular architectures or training distributions, potentially leading to biased comparisons. Neither approach fully addresses the practical needs of institutions that must process large and heterogeneous collections under real-world constraints. In practice, the relevant question is not whether a model performs best on a single curated dataset, but rather which model is most appropriate for a given collection, taking into account budget, privacy constraints, acceptable error rates, and the types of errors that can be tolerated. Our goal is to develop a benchmark designed for document recognition practitioners, enabling informed model selection across diverse, real-world conditions.

To this end, we propose a benchmark explicitly designed for document recognition practitioners. Our approach prioritizes diversity over scale, favoring a controlled but heterogeneous dataset that captures variability in scripts, languages, layouts, and document conditions. Beyond providing a dataset, we introduce a dynamic evaluation framework that produces a reproducible assessment of the state of the art at a given point in time (T1 2026), while remaining versioned and extensible. The framework is designed to integrate newly released models and increasingly challenging document subsets. This enables tracking of multilingual ATR progress instead of a one-time ranking exercise.

This paper introduces the initial version of the benchmark. We define a dataset (METATR-v1.0) and establish a standardized prompting and text normalization protocol. We also develop a comprehensive evaluation protocol that covers specialized OCR and HTR systems, open-weight OCR-oriented vision language models, large general-purpose multimodal models, and leading proprietary systems. The results are reported across complementary dimensions, including performance at the dataset and language levels, robustness with printed versus handwritten documents, computational efficiency in terms of memory consumption and inference latency, and hallucination behavior with complex documents. These analyses provide a multidimensional and practice-oriented assessment of the current state of multilingual ATR.




This paper is organised as follows: section \ref{sec:related_work} reviews existing ATR datasets and the biases they introduce into model evaluation. Section \ref{sec:benchmark} presents our benchmark, which is designed to cover a large set of languages, scripts and document types. Section \ref{sec:experiments} reports on the results of our experiments, comparing large multimodal models with specialised HTR systems and open-weight models with commercial services across the full range of benchmark challenges.

\section{Related Work}
\label{sec:related_work}


\textbf{Is HTR a solved problem?} In recent years, the field of automatic text recognition technology has grown considerably. Industry blogs and technical commentators increasingly describe OCR as a solved problem. Accuracy on typed text in high-quality images routinely exceeds 99\% on standard benchmarks, and each new frontier model is met with fresh declarations that handwriting recognition has finally been resolved too. When Gemini 3 was launched, some users claimed that it "consistently produces text with error rates comparable to the very best humans"\footnote{\url{https://generativehistory.substack.com/p/gemini-3-solves-handwriting-recognition}}.
This optimism is not entirely unfounded: modern deep learning pipelines and large multimodal models have achieved remarkable results with the types of documents that currently dominate benchmarks: clean, high-resolution scans of typed or neatly handwritten text in simple layouts and short output horizons, predominantly in English (and more generally latin scripts) or Chinese. 

However, these conditions describe only a narrow slice of the documents that ATR systems are actually asked to process. Archives, libraries and cultural heritage institutions hold vast quantities of documents that look nothing like those from IAM \cite{iam} or OCRBench \cite{ocr_bench}. These documents include degraded paper, inconsistent inks, historical scripts, multilingual content, complex multi-column layouts, non-standard page structures, and centuries of orthographic variation. When models that have been trained and evaluated using modern benchmarks are confronted with this type of documents, their performance drops sharply. \\

\textbf{Datasets: what exists and what's missing.} While many datasets exist for modern printed and handwritten documents, historical or degraded documents remain severely underrepresented, as well as most scripts and languages apart from English and Chinese.
\paragraph{Modern printed document datasets.} OCRBench \cite{ocr_bench} and olmOCR-bench are standard references for evaluating multimodal models on text recognition tasks, but they are both limited to English and Chinese. CC-OCR \cite{cc_ocr} is a notable exception in this regard, as it covers 10 languages across a diverse set of document types, although it remains focused on modern documents.
\paragraph{Modern handwriting datasets.} IAM \cite{iam} and RIMES \cite{rimes} are the standards for handwritten text recognition (HTR) evaluation, covering English and French handwritten documents, respectively. Both have been central benchmarks for over two decades and are well served by current deep learning approaches, with performance having largely plateaued. However, they are narrow in scope, covering a single language per dataset and a limited pool of writers producing clean, contemporary handwriting under controlled conditions. Consequently, they offer little insight into how models behave on real-world documents.
\paragraph{Historical document datasets.} Resources for evaluating ATR on historical documents exist, but are scarce. Although existing datasets such as READ \cite{read}, LAM \cite{lam}, POPP \cite{popp} and NorHand \cite{norhand} have contributed to improve HTR models, they are largely homogeneous in terms of script and document type. Furthermore, evaluation is usually carried out at line level rather than on full pages, avoiding the challenges of layout analysis, reading order and document-level coherence encountered in production. The release of Churro \cite{churro} addresses these shortcomings by providing the most recent and comprehensive dataset. It covers diverse scripts, languages and document conditions, and conducts evaluations at page level. Our benchmark can be seen as an extension of this effort, as described in Section 3.

\subsection{Benchmarks}

Specialized ATR systems, trained on curated document collections, have long been the dominant approach. The rise of large multimodal models has introduced a new type of general-purpose models. Comparison between the two approaches have led to mixed conclusions \cite{wolf2025-cm1,crosilla2025-benchmarkingllm}.

\textbf{Specialized ATR systems.} Traditional ATR pipelines combine document pre-processing, layout analysis and text recognition with task-specific models. Systems such as PeroOCR \cite{pero_ocr} and Azure Layout \cite{azure_layout} represent two examples of this category. PeroOCR is an open-source system designed for historical and degraded documents, while Azure Layout is a generic cloud-based service by Microsoft. Both are included in our evaluation as representatives of the specialized system category.

\textbf{Large multimodal models.} The introduction of LMMs that can process document images directly has created a new area of research in ATR. Proprietary models such as Gemini \cite{gemini}, GPT \cite{openai_gpt}, and Claude \cite{claude_opus}, as well as OCR-specialised open-weight models such as olmOCR \cite{olm_ocr} and LightOnOCR \cite{lighton_ocr} are being used more and more for tasks that go beyond what they were originally designed for. Churro \cite{churro} zero-shot evaluation provides an insight on the current state: among proprietary models, Gemini 2.5 Pro achieves the strongest performance at 80.9\%, while OCR-specialised open-weight models cluster around 70\%. But even the best models often have problems with historical documents. The authors found that more than a third of predictions from a small open-weight model had major hallucinations, and over 40\% had reading-order errors on multi-column pages.

\begin{figure}[t!]
    \centering
    \includegraphics[width=\linewidth]{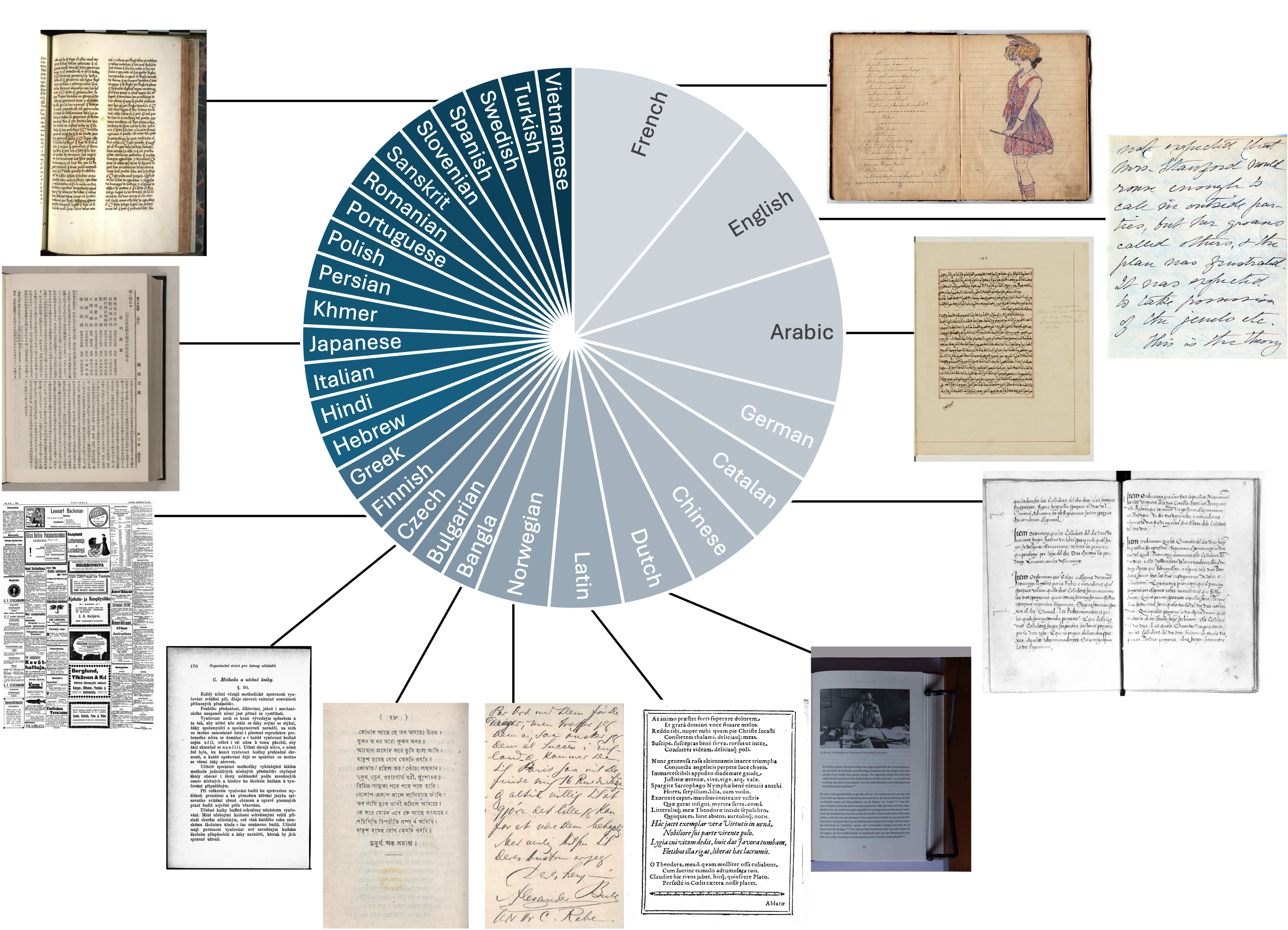}
    \caption{Distribution of languages and example images in the benchmark dataset.}
    \label{fig:dataset_details_examples}
\end{figure}

\textbf{LMMs vs. specialized systems: a fragmented picture.} Direct comparisons between LMMs and specialized HTR systems consistently reveal that no single model dominates across all conditions. On modern English handwriting, LMMs prove competitive, but specialized systems retain a clear advantage on German, multilingual, and historical documents \cite{crosilla2025-benchmarkingllm}.
Interestingly, LMMs have been shown to outperform specialized systems in zero-shot settings, while the situation reverses in few-shot scenarios \cite{wolf2025-cm1}. A further complication is the evaluation protocol itself: most existing comparisons operate at the line level, where layout and reading order are given. Churro's page-level evaluation makes this explicit: even the strongest proprietary model reaches only 80.9\% accuracy on historical documents in zero-shot conditions. Most models still struggle with reading-order errors and hallucinations \cite{churro}.

\section{Multilingual Real-World Benchmark}
\label{sec:benchmark}

Our primary objective in building an evaluation dataset for LLMs and HTR models was to maximize document diversity rather than quantity. The resulting benchmark, which we define as METATR v1.0\footnote{METATR v1.0: \url{https://huggingface.co/datasets/Teklia/ATR-benchmark}}, comprises 453 images, including full pages, double-pages, and isolated paragraphs. These images were manually selected from test sets extracted from 17 existing datasets and cover 29 languages. A significant portion of the benchmark comes from the Churro database \cite{churro}, from which we sampled ten pages per language, carefully selecting both handwritten and printed documents. The choice of ten pages per language is a deliberate trade-off. This number is generally sufficient to capture meaningful variability in layout, script, and document condition within each dataset, while remaining small enough to keep the final benchmark at a reasonable and reproducible scale. Overall, the dataset contains 63\% handwritten documents and 37\% printed documents. This distribution ensures a strong focus on challenging HTR scenarios while maintaining diversity in printed documents.

\input{tables/dataset_details}

The selected languages and datasets were not chosen arbitrarily. Most are widely used by the research community, facilitating comparison with prior work while reflecting the linguistic and documentary reality of many ongoing digitization projects. Together, they represent a broad spectrum of scripts and institutional use cases, including administrative records, personal correspondence, newspapers, literary texts, and religious manuscripts.

When developing METATR (v1.0), we intentionally included various document types, languages, and layouts. These include single-page, double-page, and multi-column formats, such as newspapers, with or without illustrations. These document types correspond to specific institutional needs: newspapers and printed books reflect mass digitization and searchability projects, while handwritten letters and administrative registers correspond to archival transcription projects. Historical manuscripts represent scholarly and cultural heritage use cases. While the benchmark captures only a fraction of the vast diversity of historical and modern documents, it provides a representative sample of the types of documents most frequently encountered in current digitization workflows.

While the Churro database is a valuable resource, its size makes it difficult to use directly for controlled evaluation. Therefore, we favored a curated and diverse subset over exhaustive coverage, enabling fair and repeatable comparisons. Finally, we integrated reference datasets widely used in the literature for model evaluation, allowing newly developed systems to be directly compared with the models assessed in our study. Figure \ref{fig:dataset_details_examples} illustrates the language distribution of the dataset and provides representative example images, while Table \ref{tab:dataset_details} details its full composition.

This benchmark is designed as a dynamic, evolving resource. Future versions will incorporate new projects, newly available datasets, and emerging document types of interest. Thus, METATR is intended as a dynamic evaluation framework that can adapt to ongoing developments in ATR research and real-world digitization needs, rather than a static resource.

\section{Experiments and Results}
\label{sec:experiments}


\subsection{Models}

The models included in our evaluation were selected according to several practical and scientific criteria. We focused on systems that are easy to test and reproduce, such as those available through standard libraries like Hugging Face or accessible APIs. We also selected systems that can be used with reasonable computational resources, no more than a single A100 GPU for open-source models, and that produce strong results on existing benchmarks. Additionally, we focused on models that are frequently cited or widely adopted in the community to ensure that our comparison remains relevant and informative for future research.

\textbf{Specialized OCR and HTR systems.} First, we evaluated dedicated OCR systems, which are optimized specifically for text detection and transcription. We considered Microsoft Azure Layout \cite{azure_layout}, which is a structured document analysis system capable of extracting text while preserving layout information; and Microsoft Azure OCR \cite{azure_ocr}. However, since the Azure OCR model produced lower results than the Azure Layout model, we do not report its results in the paper. We also included PeroOCR \cite{pero_ocr}, a widely used academic OCR framework for historical and handwritten document recognition. These systems provide strong baselines based on traditional OCR pipelines.

\textbf{Open-source vLLMs.} We evaluated several open-source vision-language models (vLLMs), which process visual and textual information jointly and can generate transcriptions directly from images. These models include Churro (3B) \cite{churro}, a multilingual, OCR-oriented model; DeepSeek-OCR (3B) \cite{deepseek_ocr} and DeepSeek-OCR-2 (3B) \cite{deepseek_ocr2}, which are designed for end-to-end document understanding; LightOnOCR-2 (1B) \cite{lighton_ocr}, a lightweight model optimized for efficient deployment; olmOCR-2 (8B) \cite{olm_ocr}; Qwen3-VL (8B) \cite{qwen3_vl}, a large, multimodal language model; and RolmOCR (8B) \cite{rolm_ocr}, another large-scale, OCR-oriented model. These models represent a diverse range of parameter sizes (from 1 billion to 8 billion) and architectural trade-offs between efficiency and performance.

\textbf{Closed-source VLLMs.} Finally, we included several closed-source vLLMs to compare open approaches with state-of-the-art, proprietary systems. These include Claude Opus 4.5 \cite{claude_opus}, Gemini 3 Pro \cite{gemini}, Mistral AI Small, Medium, and Large \cite{mistral_small,mistral_medium,mistral_large}, Mistral OCR \cite{mistral_ocr}, and GPT-5.1 \cite{openai_gpt}. These proprietary systems usually have access to large-scale training data and substantial computational resources, which allows them to often define the current performance standard for multimodal document understanding. Including them allows us to position open-source models in comparison to leading commercial solutions.

In the future, we aim to expand this benchmark and evaluate new models on more recent versions of the dataset. The models tested will have to meet the same conditions as those in this article, especially regarding easy access and reasonable memory usage.

\subsection{Evaluation settings}

All models were evaluated under consistent prompting conditions. For models that were explicitly fine-tuned for OCR tasks, we used their default recommended prompt. For all other vision-language models, we applied a single, generic instruction: "Extract the text from this image. Do not add any context." This instruction was designed to minimize the effects of prompt engineering and ensure comparability across systems. We kept model hyperparameters at their default values (e.g., temperature and maximum number of tokens) to reflect standard usage. The only exception was for long or densely written pages, for which we increased the maximum token limit to ensure the entire page's content could be generated without truncation. For evaluation purposes, both the predictions and the ground-truth transcriptions were normalized using the same preprocessing pipeline: text was lowercased, and line breaks were replaced with spaces. This normalization ensured a fair and consistent computation of the Character Error Rate (CER) across models.

\input{tables/cer_by_dataset}

\subsection{Evaluation results}

\subsubsection{Results by dataset.}

Table \ref{tab:results_dataset} reports the CER obtained by each model across all datasets in our benchmark. The results highlight three main observations: (1) the strong performance of some recent proprietary vLLMs, (2) the competitive performance of some open-source vLLMs, and (3) the robustness, but limited multilingual generalization, of traditional OCR systems.

First, proprietary multimodal models show contrasted performance levels. Gemini 3 Pro and Claude Opus 4.5 clearly stand out, consistently achieving the lowest CER across a wide range of datasets, including challenging scripts such as Arabic, Japanese, and Khmer. In contrast, the Mistral models perform significantly worse. They often rank well behind Gemini and Claude and, in several cases, behind some large open-source models, such as Churro and RolmOCR. These results suggest that not all proprietary systems benefit equally from large-scale multimodal pretraining and that model architecture and training data composition play crucial roles in multilingual document understanding performance.

Second, performance is heterogeneous among open-source vLLMs. While smaller models, such as LightOnOCR-2 and DeepSeek-OCR, struggle with complex layouts or non-Latin scripts, larger models, such as QWEN3-VL and olmOCR-2, demonstrate competitive results on several datasets, including FINLAM, IAM, and subsets of Churro. In addition, the Churro model, despite being much smaller, achieves competitive results compared to the proprietary Claude and Gemini models. This is mainly because it was trained on the Churro dataset, which is highly similar to our test data. However, open-source models also exhibit higher variance, occasionally producing extreme CER values for certain languages (e.g., Japanese, Khmer, and Greek), which indicates instability in multilingual generalization.

Third, traditional OCR systems present an interesting contrast. Azure Layout performs well on clean documents (e.g., IAM and RIMES), sometimes outperforming several vLLMs. However, its performance decreases on more challenging multilingual or handwritten datasets. PeroOCR excels in specific European languages, such as Czech and Slovenian, but struggles with many non-Latin scripts. These results confirm that classical OCR pipelines remain competitive in controlled settings but are less adaptable to diverse, multilingual benchmarks.

At the dataset level, results clearly show that handwritten and historical collections (e.g., READ-2016, ScribbleLens, and Esposalles) remain particularly challenging for most systems, with substantial performance gaps between models. In contrast, more standardized modern datasets (e.g., FINLAM, IAM and RIMES) yield relatively low CER across multiple approaches. In summary, while large-scale, proprietary multimodal models currently define the upper limit of performance, several open-source models are competitive alternatives under realistic computational constraints. At the same time, variability in performance across scripts and layouts confirms the importance of our diverse benchmark for thoroughly assessing multilingual document understanding systems.

\input{tables/cer_by_language}

\subsubsection{Results by language.}

Table \ref{tab:results_language} complements the dataset-level analysis by reporting CER aggregated per language. Overall, the trends remain consistent with those in Table \ref{tab:results_dataset}. Gemini 3 Pro and Claude Opus 4.5 dominate most scripts, with Gemini achieving the best CER in many languages, including Arabic, French, Japanese, and Chinese. Claude is also highly competitive, occasionally outperforming Gemini (e.g. in Spanish and Norwegian), which confirms the superiority of these two models in multilingual settings. The Mistral models lag significantly behind Gemini and Claude and are often outperformed by large open-source models, particularly on complex or non-Latin scripts. Among open-source systems, larger models, such as QWEN3-VL and olmOCR-2, deliver competitive results in several high-resource languages. Meanwhile, Churro achieves strong performance in languages similar to its training data. However, open-source models exhibit higher variance. Traditional OCR systems such as Azure Layout and PeroOCR perform well in specific languages (e.g., Arabic, Czech, and Slovenian), but they lack consistent cross-lingual robustness. Overall, the language-level results reinforce the conclusion that large, proprietary, multimodal models currently provide the most consistent multilingual performance.

\input{tables/cer_handwritten_printed}

\subsubsection{Results on printed and handwritten documents.}

Table \ref{tab:results_printed_handwritten} refines the analysis further by separating printed and handwritten documents. The results confirm the observed trends: Gemini 3 Pro outperforms all other models in both settings, achieving the best CER on printed documents (17.6\%) and handwritten documents (28.5\%), while Claude Opus 4.5 ranks second on handwritten documents (31.2\%). Azure Layout obtains the second-best performance on printed documents (30.8\%), highlighting the strength of traditional OCR pipelines on well-structured documents. In contrast, performance gaps widen considerably on handwritten documents, where most open-source models, especially smaller ones, such as LightOnOCR-2 and DeepSeek-OCR, show substantial degradation. Larger open-source models, such as QWEN3-VL, remain competitive but still lag behind top proprietary systems. Mistral models again perform significantly worse, especially on handwritten data. These results reinforce the conclusion that handwritten text remains the most challenging scenario and that large, proprietary, multimodal models currently provide the most robust performance across document types.

\input{tables/dataset_computational_evaluation}

\subsection{Computational efficiency}

To evaluate ATR models in real-world conditions, practical deployment constraints must be considered along with model accuracy. To this end, we assessed all models along two metrics: GPU memory consumption and inference time.
These measurements were conducted on a selection of five representative documents spanning a broad range of complexity: a simple paragraph; a single-column page; a paragraph in a low-resource language; a multi-column page; and a long page with a complex reading order. This selection contrasts easy, short inputs with challenging, large ones, enabling us to capture both best- and worst-case computational behavior. A detailed summary of these documents and their main characteristics is provided in Table \ref{tab:dataset_comp_eval}.

Open-source models were benchmarked on a single NVIDIA A100 GPU using the Hugging Face \texttt{transformers} library with Flash Attention 2 and a batch size of 1, while API-based models were queried via paid accounts under standard conditions. All the results for this analysis are summarized in Table \ref{tab:comp_eval}.
Among locally run models, PeroOCR is the fastest (0.25s/100w) due to its lightweight architecture. It is also the most memory-efficient, using less than 1 GiB. For vision-language models, the time taken for inference scales with model size: 7–9 GiB and 4.5–8.5 seconds per 100 words for 2–3 billion parameter models, rising to around 20 GiB for 7–8 billion parameter models. However, there are two notable exceptions: olmOCR-2 and Churro are significantly slower than comparable models. Both models predict structured markup alongside the transcription, YAML for olmOCR-2 and XML for Churro, which substantially increases the number of generated tokens.

\input{tables/computational_evaluation}

API-based models introduce additional sources of variance that make direct comparisons with locally hosted systems more complicated. Response times are affected by server-side factors (queuing, resource allocation and network latency), as well as provider-specific steps (image resizing, meta prompts, post-processing) that are not always disclosed. Features such as thinking modes, which were used by Gemini in our evaluation, can further increase latency. 
Among API-based models, Mistral OCR and Azure Layout are the fastest (<1s/100w), which can be explained by their document-specialized architectures. General vision–language APIs, however, vary between 2 to 10s per 100 words. Gemini is the slowest (10.31s/100w), partly due to its extended thinking mode. Mistral AI's models show a clear gradient of speed and capability from small (2.13s/100w) to large (5.30s/100w).

When considered together, the accuracy and computational analysis reveal a clear trade-off between performance and deployment cost. Large, proprietary models, especially Gemini 3 Pro and Claude Opus 4.5, consistently deliver the lowest CER. However, this superior accuracy comes at the cost of higher latency, as well as monetary expense and limited transparency regarding inference pipelines. Conversely, traditional OCR systems, such as Azure Layout, and lightweight models, such as PeroOCR, offer extremely fast inference and minimal resource consumption. This makes them attractive for large-scale or time-sensitive applications. However, they have reduced robustness in multilingual and handwritten settings. Open-source vision-language models occupy an intermediate position. While larger models can approach the performance of proprietary models in certain scenarios, they require substantial GPU memory (up to 20 GiB) and exhibit greater variability, especially with complex scripts. Overall, these results suggest that model selection should depend on the intended deployment context. Proprietary large-scale models currently maximize accuracy and robustness. However, lighter OCR systems and smaller open-source models are more efficient and cost-effective alternatives when computational constraints are critical.

\section{Conclusion}

This work introduced METATR (v1.0), a multilingual, real-world benchmark designed to evaluate automatic text recognition systems under diverse, practical conditions. Prioritizing diversity over scale, we created an heterogeneous dataset covering 29 languages, multiple scripts, and varied layouts documents. This approach moves beyond the narrow evaluation settings that dominate many current benchmarks, providing a more realistic assessment of ATR performance in archival and cultural heritage contexts.

Our experimental results demonstrate that large, proprietary, multimodal models, particularly Gemini 3 Pro and Claude Opus 4.5, currently represent the upper limit of performance across most datasets, languages, and document types. However, performance remains highly dependent on script and layout complexity. Open-source vision-language models demonstrate competitive results in several scenarios, especially as model size increases. However, they exhibit higher variance and reduced robustness with complex or low-resource scripts. Traditional OCR systems are highly efficient on well-structured printed documents but struggle with multilingual and historical pages.

Taken together, these findings challenge the notion that ATR is a solved problem. Although significant progress has been made, achieving robust multilingual document understanding at the page level, particularly for historical documents, remains a significant challenge. By combining accuracy, robustness, and computational analysis, METATR provides a multidimensional, practitioner-oriented framework for selecting models in real-world deployment scenarios. This version of the dataset, along with details of the model versions used and the evaluation code, will be made publicly available.

\section{Future Work}

METATR is designed to be a dynamic benchmark rather than a static dataset. Several directions will guide its future development. First, we plan to evaluate new systems using the same standardized prompt, normalization, and reporting conditions. This will ensure the reproducibility and comparability of results, as well as transparency regarding model versions and inference parameters.

Second, we will update the datasets as new digitization projects emerge and new datasets become publicly available. Selected subsets will be incorporated into future versions. Each update will be versioned and documented to maintain comparability while gradually increasing diversity and difficulty.

Third, to mitigate the risk of overfitting benchmarks, we plan to introduce a private test set. This would allow public development on a visible subset while reserving a confidential portion for official evaluation. This approach encourages true generalization rather than optimization for a fixed and fully exposed dataset.

Finally, we plan to regularly publish state-of-the-art reports based on evolving benchmarks. These reports will provide progress updates on ATR for different languages, scripts, and document types. This will enable us to track improvements and identify remaining challenges. Through these efforts, METATR is designed to evolve alongside the research field while providing a stable but adaptable reference framework for real-world, multilingual ATR evaluation.




\bibliographystyle{splncs04}
\bibliography{refs}

\end{document}

%% file: tables/dataset_details.tex
\begin{table}[t!]
    \centering
    \fontsize{6.5pt}{6.5pt}\selectfont
    \setlength{\tabcolsep}{2pt}
    \caption{Overview of the METATR (v1.0) dataset composition by source, language, and document type.}
    \label{tab:dataset_details}
    \begin{tabular}{l|clllll}
    \hline
        \textbf{Dataset} & \textbf{Images} & \textbf{Level} & \textbf{Language(s)} & \textbf{Date} & \textbf{Writing} & \textbf{Type} \\ \hline
        BnL\tablefootnote{BNL Starter pack: \url{https://data.bnl.lu/data/historical-newspapers/}} & 3 & Paragraph & German & 1868 / 1877 & Printed & Historical newspapers \\ \hline
        CASIA- & \multirow{2}{*}{10} & \multirow{2}{*}{Page} & \multirow{2}{*}{Chinese} & \multirow{2}{*}{2007-2010} & \multirow{2}{*}{Manuscript} & \multirow{2}{*}{News} \\
        HWDB2.0-2.2 \cite{casia} & \\ \hline
        Churro \cite{churro} & 290 & Page & Multilingual & - & Both & Mix \\ \hline
        \multirow{2}{*}{DAI-CRETDHI\tablefootnote{\href{https://dai-cretdhi.univ-lr.fr/}{DAI-CRETDHI}: \url{https://huggingface.co/datasets/Teklia/DAI-records-ATR}}} & \multirow{2}{*}{10} & \multirow{2}{*}{Paragraph} & \multirow{2}{*}{French} & \multirow{2}{*}{1600-1800} & \multirow{2}{*}{Manuscript} & Parish and \\
        & & & & & & civil records \\ \hline
        DIY History\tablefootnote{DIY History Social Justice: \url{http://diyhistory.lib.uiowa.edu/}} & 20 & Page & English & - & Manuscript & Mix \\ \hline
        \multirow{2}{*}{Esposalles \cite{esposalles}} & \multirow{2}{*}{10} & \multirow{2}{*}{Paragraph} & \multirow{2}{*}{Spanish} & \multirow{2}{*}{17th c.} & \multirow{2}{*}{Manuscript} & Parish marriage \\
        & & & & & & records \\ \hline
        \multirow{2}{*}{FINLAM\tablefootnote{\href{https://projets.litislab.fr/finlam/}{FINLAM}: \url{https://huggingface.co/datasets/Teklia/Newspapers-finlam}}} & \multirow{2}{*}{10} & \multirow{2}{*}{Paragraph} & English + & \multirow{2}{*}{1925-1930} & \multirow{2}{*}{Printed} & \multirow{2}{*}{Historical newspapers} \\
        & & & French & & \\ \hline
        HORAE \cite{horae} & 10 & Page & Latin & Middle Ages & Manuscript & Books of hours \\ \hline
        IAM \cite{iam} & 10 & Paragraph & English & Modern & Manuscript & Letters \\ \hline
        \multirow{2}{*}{NorHand (v3) \cite{norhand}} & \multirow{2}{*}{10} & \multirow{2}{*}{Paragraph} & \multirow{2}{*}{Norwegian} & \multirow{2}{*}{19th-20th c.} & \multirow{2}{*}{Manuscript} & Letter and \\ 
        & & & & & & diary documents \\ \hline
        OpenITI \cite{openiti} & 10 & Page & Arabic & 16th c. & Printed & Literary texts \\ \hline
        PELLET\tablefootnote{Pellet: \url{https://europeana.transcribathon.eu/documents/story/?story=121795}} & 10 & Page & French & 1914-1918 & Manuscript & Personal writings \\ \hline
        QARI\tablefootnote{QARI: \url{https://huggingface.co/datasets/oddadmix/qari-0.2.2-news-dataset-large}} & 10 & Page & Arabic & Modern & Printed & News articles \\ \hline
        RASM \cite{rasm} & 10 & Page & Arabic & 9th-19th c. & Manuscript & Scientific manuscripts \\ \hline
        READ-2016 \cite{read} & 10 & Page & German & 1470-1805 & Manuscript & Administrative records \\ \hline
        RIMES \cite{rimes} & 10 & Page & French & Modern & Manuscript & Letters \\ \hline
        \multirow{2}{*}{ScribbleLens \cite{scribblelens}} & \multirow{2}{*}{10} & \multirow{2}{*}{Page} & \multirow{2}{*}{Dutch} & \multirow{2}{*}{16th-18th c.} & \multirow{2}{*}{Manuscript} & Ship journals and \\
        & & & & & & daily logbooks \\ \hline
    \end{tabular}
\end{table}

%% file: tables/cer_by_dataset.tex
\begin{table}[ht!]
    \centering
    \fontsize{6.5pt}{6.5pt}\selectfont
    \setlength{\tabcolsep}{1.5pt}
    \caption{Character Error Rate (\%) for each dataset and model. Best values are highlighted in bold and second-best results are underlined.}
    \label{tab:results_dataset}
    \begin{tabular}{l|rrrrrrr|rr|rrrrrrr}
    \hline
        ~ & \rotatebox[origin=c]{90}{\textbf{\shortstack[c]{Churro\\(3B)}}} & \rotatebox[origin=c]{90}{\textbf{\shortstack[c]{DeepSeek-\\OCR (3B)}}} & \rotatebox[origin=c]{90}{\textbf{\shortstack[c]{DeepSeek-\\OCR-2 (3B)}}} & \rotatebox[origin=c]{90}{\textbf{\shortstack[c]{LightOn\\OCR-2 (1B)}}} & \rotatebox[origin=c]{90}{\textbf{\shortstack[c]{olmOCR-2\\(8B)}}} & \rotatebox[origin=c]{90}{\textbf{\shortstack[c]{QWEN3-VL\\(8B)}}} & \rotatebox[origin=c]{90}{\textbf{\shortstack[c]{RolmOCR\\(8B)}}} & \rotatebox[origin=c]{90}{\textbf{\shortstack[c]{Azure\\Layout}}} & \rotatebox[origin=c]{90}{\textbf{PeroOCR}} & \rotatebox[origin=c]{90}{\textbf{\shortstack[c]{Claude\\Opus 4.5}}} & \rotatebox[origin=c]{90}{\textbf{\shortstack[c]{Gemini\\3 Pro}}} & \rotatebox[origin=c]{90}{\textbf{GPT-5.1}} & \rotatebox[origin=c]{90}{\textbf{\shortstack[c]{Mistral AI\\small}}} & \rotatebox[origin=c]{90}{\textbf{\shortstack[c]{Mistral AI\\medium}}} & \rotatebox[origin=c]{90}{\textbf{\shortstack[c]{Mistral AI\\large}}} & \rotatebox[origin=c]{90}{\textbf{\shortstack[c]{Mistral\\OCR}}} \\ \hline
        \textbf{BnL} & 9.6 & 36.3 & 67.1 & 248.9 & 6.5 & 6.3 & 6.4 & 32.0 & \underline{6.1} & 16.1 & \textbf{5.7} & 39.2 & 81.0 & 45.6 & 51.4 & 73.4 \\
        \textbf{CASIA} & 106.9 & 23.4 & 53.7 & 806.0 & 10.2 & \underline{8.6} & 11.5 & 12.3 & 179.1 & 13.8 & \textbf{5.0} & 41.8 & 904.3 & 72.5 & 72.9 & 52.4 \\
        \textbf{Ch-Arabic} & 103.1 & 361.4 & 80.8 & 218.6 & 55.6 & 76.1 & 73.6 & \underline{31.6} & 97.3 & 36.7 & \textbf{21.0} & 77.1 & 101.1 & 91.6 & 96.8 & 32.9 \\
        \textbf{Ch-Bangla} & 17.6 & 34.2 & 53.3 & 87.8 & 15.7 & 23.4 & 19.9 & 74.0 & 82.3 & \underline{8.8} & \textbf{6.5} & 20.8 & 25.0 & 21.2 & 20.8 & 23.0 \\
        \textbf{Ch-Bulgarian} & \underline{7.2} & 36.2 & 43.9 & 18.8 & 11.4 & 7.9 & 24.6 & 10.8 & 13.1 & 31.4 & \textbf{4.6} & 37.0 & 32.2 & 28.4 & 23.6 & 34.5 \\
        \textbf{Ch-Catalan} & \textbf{8.8} & 162.3 & 68.6 & 132.2 & 15.3 & 18.7 & 18.5 & 32.0 & 47.3 & \underline{10.1} & 16.2 & 29.9 & 20.7 & 19.6 & 20.4 & 19.9 \\
        \textbf{Ch-Chinese} & 188.4 & 155.9 & 181.8 & 219.8 & 119.3 & 95.2 & 110.4 & \underline{90.8} & 98.9 & 92.4 & \textbf{83.5} & 97.4 & 114.6 & 98.7 & 110.3 & 96.5 \\
        \textbf{Ch-Czech} & 6.5 & 2.8 & 4.1 & 1.7 & 2.0 & 1.4 & 2.1 & 2.8 & \textbf{0.9} & 1.4 & \underline{1.1} & 6.6 & 2.8 & 3.1 & 2.8 & 1.6 \\
        \textbf{Ch-Dutch} & \underline{3.7} & 33.2 & 51.1 & 53.5 & 21.2 & 5.4 & 41.9 & 7.5 & 27.6 & 51.3 & \textbf{2.3} & 23.8 & 52.4 & 51.5 & 45.7 & 49.4 \\
        \textbf{Ch-English} & 64.2 & 91.3 & 72.1 & 77.9 & 47.5 & \underline{36.4} & 69.1 & 39.7 & 83.9 & 73.7 & \textbf{23.3} & 80.1 & 82.1 & 74.6 & 71.3 & 82.1 \\
        \textbf{Ch-Finnish} & 56.6 & 88.2 & 88.1 & 84.7 & 55.9 & 50.2 & 76.0 & \underline{32.2} & 52.9 & 75.8 & \textbf{19.4} & 80.9 & 87.7 & 82.4 & 77.0 & 75.2 \\
        \textbf{Ch-French} & 24.3 & 23.3 & 23.1 & 45.3 & \underline{9.4} & 11.9 & 37.2 & 14.7 & 16.6 & 21.9 & \textbf{8.0} & 15.9 & 15.6 & 17.8 & 13.4 & 12.2 \\
        \textbf{Ch-German} & 62.3 & 130.1 & 120.2 & 137.8 & \underline{53.4} & 63.8 & 66.0 & 75.5 & 68.2 & 59.5 & \textbf{34.8} & 69.6 & 62.7 & 70.8 & 70.2 & 65.9 \\
        \textbf{Ch-Greek} & \textbf{24.2} & 662.5 & 126.1 & 439.0 & 179.8 & 65.3 & 134.3 & 79.6 & 90.6 & 47.0 & \underline{45.8} & 72.9 & 524.4 & 78.3 & 91.6 & 84.2 \\
        \textbf{Ch-Hebrew} & 239.4 & 178.9 & 89.2 & 197.5 & 115.1 & 89.2 & 128.0 & 106.9 & 93.7 & \underline{71.4} & \textbf{52.0} & 93.6 & 226.1 & 89.6 & 77.5 & 73.5 \\
        \textbf{Ch-Hindi} & \underline{8.2} & 24.3 & 62.8 & 198.2 & 17.1 & 15.2 & 17.5 & 32.4 & 94.4 & 14.3 & \textbf{7.8} & 32.3 & 19.2 & 20.8 & 48.6 & 33.7 \\
        \textbf{Ch-Italian} & 16.6 & 44.7 & 71.6 & 23.4 & 14.0 & 13.2 & 17.1 & 27.2 & 53.9 & \textbf{10.3} & \underline{10.4} & 39.7 & 21.5 & 20.3 & 24.0 & 16.2 \\
        \textbf{Ch-Japanese} & \underline{43.6} & 863.0 & 255.2 & 1011.9 & 403.6 & 64.2 & 197.1 & 51.9 & 102.8 & 62.3 & \textbf{35.2} & 90.9 & 100.4 & 126.0 & 223.7 & 62.4 \\
        \textbf{Ch-Khmer} & \underline{84.5} & 559.8 & 94.5 & 1119.7 & 284.7 & 122.9 & 259.7 & 96.7 & 99.0 & 103.9 & \textbf{67.2} & 98.6 & 129.4 & 100.6 & 113.9 & 96.1 \\
        \textbf{Ch-Latin} & \textbf{15.6} & 42.9 & 38.0 & 57.7 & 24.9 & 24.9 & 29.7 & 29.6 & 29.2 & 19.9 & \underline{16.4} & 43.9 & 43.8 & 28.7 & 32.0 & 36.3 \\
        \textbf{Ch-Norwegian} & \underline{16.7} & 83.6 & 164.4 & 43.1 & 24.4 & 57.9 & 28.4 & 43.0 & 76.1 & \textbf{12.9} & \underline{16.7} & 37.4 & 49.1 & 50.7 & 88.1 & 21.7 \\
        \textbf{Ch-Persian} & 206.5 & 321.9 & 106.5 & 192.8 & 180.7 & 56.2 & 47.8 & 55.1 & 97.7 & \underline{29.7} & \textbf{21.9} & 58.4 & 622.2 & 64.9 & 69.2 & 42.4 \\
        \textbf{Ch-Polish} & 28.7 & 89.8 & 89.7 & 59.9 & 39.1 & \underline{23.2} & 54.3 & 37.4 & 64.0 & 55.3 & \textbf{17.8} & 64.1 & 71.8 & 65.8 & 62.2 & 64.7 \\
        \textbf{Ch-Portuguese} & 81.6 & 221.2 & 127.6 & 280.8 & 72.8 & 88.1 & 156.5 & 68.6 & 73.9 & \textbf{55.6} & \underline{59.1} & 84.4 & 1346.9 & 82.1 & 91.5 & 80.2 \\
        \textbf{Ch-Romanian} & \textbf{21.1} & 82.4 & 104.7 & 41.6 & 50.1 & 40.7 & 38.9 & 45.9 & 48.5 & \underline{32.1} & 41.3 & 67.9 & 46.4 & 39.7 & 54.2 & 40.5 \\
        \textbf{Ch-Sanskrit} & \textbf{39.1} & 238.7 & 224.7 & 188.0 & 190.5 & 50.8 & 72.3 & 46.6 & 95.3 & 43.6 & \underline{39.9} & 62.8 & 55.5 & 53.3 & 63.7 & 50.3 \\
        \textbf{Ch-Slovenian} & \underline{1.7} & 5.0 & 6.5 & 4.3 & 4.6 & 2.9 & 4.4 & 5.1 & \textbf{1.4} & 3.6 & \underline{1.7} & 4.7 & 3.6 & 4.9 & 4.0 & 5.4 \\
        \textbf{Ch-Spanish} & 24.1 & 179.6 & 76.2 & 153.0 & 30.8 & 39.5 & 85.1 & 26.0 & 41.4 & \textbf{14.4} & \underline{15.3} & 39.3 & 44.8 & 48.2 & 42.2 & 27.3 \\
        \textbf{Ch-Swedish} & 49.1 & 102.2 & 90.3 & 68.2 & 49.5 & \underline{29.2} & 60.1 & 44.8 & 71.6 & 52.4 & \textbf{18.2} & 56.5 & 68.5 & 63.0 & 61.9 & 62.0 \\
        \textbf{Ch-Turkish} & \textbf{29.7} & 270.1 & 77.3 & 198.3 & 585.9 & 112.4 & 172.1 & 60.4 & 96.6 & 59.3 & 63.6 & 69.5 & 98.0 & 106.3 & 97.8 & \underline{54.7} \\
        \textbf{Ch-Vietnamese} & \textbf{40.4} & 59.0 & 144.7 & 1666.2 & 49.9 & 52.7 & 46.3 & 57.6 & 100.8 & 77.0 & \underline{43.1} & 84.8 & 1098.8 & 119.7 & 142.3 & 86.4 \\
        \textbf{DAI-CRETDHI} & 18.8 & 316.0 & 96.8 & 699.9 & 32.0 & 91.8 & 38.5 & 41.6 & 71.3 & \underline{13.0} & \textbf{10.2} & 36.9 & 56.6 & 51.4 & 64.6 & 28.7 \\
        \textbf{DIY History} & 16.7 & 20.0 & 23.8 & 51.3 & 4.6 & 14.8 & 38.2 & 9.3 & 33.1 & \textbf{3.6} & \underline{4.3} & 12.9 & 590.2 & 10.0 & 16.3 & 4.7 \\
        \textbf{Esposalles} & 16.8 & 899.3 & 52.0 & 1693.4 & \underline{15.1} & 22.2 & \textbf{14.1} & 32.8 & 43.9 & 18.5 & 20.6 & 15.2 & 22.0 & 26.7 & 43.8 & 78.8 \\
        \textbf{FINLAM} & 2.2 & 1.6 & 4.9 & 1.6 & 0.8 & \textbf{0.3} & 7.2 & 2.4 & 1.2 & \underline{0.4} & 0.5 & 2.5 & 2.0 & 0.7 & 1.6 & 1.5 \\
        \textbf{HORAE} & 24.7 & 69.1 & 73.8 & 388.0 & 30.6 & 34.9 & 37.0 & 49.7 & 57.5 & \underline{11.7} & \textbf{7.4} & 52.6 & 64.7 & 35.2 & 58.5 & 61.2 \\
        \textbf{IAM} & 6.5 & 11.3 & 18.1 & 5.8 & 5.4 & 4.8 & 5.2 & \textbf{3.9} & 67.6 & 4.4 & \underline{4.0} & 5.5 & 5.3 & 4.1 & 8.5 & \underline{4.0} \\
        \textbf{NorHand} & 16.1 & 81.1 & 77.5 & 33.2 & 12.4 & 28.1 & 17.2 & 20.6 & 56.8 & \underline{10.2} & \textbf{8.3} & 24.9 & 30.2 & 26.0 & 38.2 & 14.1 \\
        \textbf{OpenITI} & 13.2 & 180.6 & 31.1 & 17.5 & 12.9 & 12.4 & 15.1 & 10.5 & 93.9 & \textbf{10.0} & \underline{10.1} & 36.2 & 37.1 & 40.1 & 45.8 & 16.0 \\
        \textbf{PELLET} & 17.5 & 401.0 & 63.7 & 53.3 & 18.1 & 18.0 & 148.0 & 29.3 & 78.1 & \underline{12.3} & \textbf{5.4} & 25.8 & 24.7 & 21.8 & 31.7 & 16.4 \\
        \textbf{QARI} & 19.2 & 5.3 & 47.3 & 13.2 & 12.7 & 27.6 & 8.3 & 2.7 & 92.1 & 3.9 & \underline{2.4} & 43.0 & 62.3 & 46.5 & 52.0 & \textbf{2.3} \\
        \textbf{RASM} & 29.4 & 220.4 & 77.7 & 159.8 & 119.5 & 43.5 & 37.8 & 27.1 & 96.8 & \textbf{16.2} & \underline{23.7} & 101.0 & 522.7 & 81.8 & 87.9 & 27.0 \\
        \textbf{READ-2016} & \textbf{15.4} & 673.2 & 165.5 & 560.9 & 599.1 & 258.1 & 507.3 & 72.9 & 84.7 & \underline{48.8} & 56.3 & 72.8 & 141.1 & 193.3 & 84.3 & 72.3 \\
        \textbf{RIMES} & 7.5 & 10.2 & 13.7 & 33.3 & 5.5 & 6.5 & 16.9 & \textbf{2.6} & 52.8 & \underline{2.7} & 3.9 & 3.5 & 6.8 & 8.1 & 9.0 & 6.1 \\
        \textbf{ScribbleLens} & \textbf{15.7} & 399.5 & 110.3 & 288.5 & 43.9 & 55.1 & 105.3 & 53.4 & 76.0 & \underline{27.8} & 49.7 & 77.4 & 66.1 & 50.8 & 68.5 & 47.8 \\ \hline
        \textbf{Average rank} & 3 & 16 & 14 & 15 & 6 & 4 & 8 & 5 & 12 & 2 & 1 & 9 & 13 & 10 & 11 & 7 \\ \hline
    \end{tabular}
\end{table}

%% file: tables/cer_by_language.tex
\begin{table}[ht!]
    \centering
    \fontsize{6.5pt}{6.5pt}\selectfont
    \setlength{\tabcolsep}{2pt}
    \caption{Character Error Rate (\%) for each language and model. Best values are highlighted in bold and second-best results are underlined.}
    \label{tab:results_language}
    \begin{tabular}{l|rrrrrrr|rr|rrrrrrr}
    \hline
        ~ & \rotatebox[origin=c]{90}{\textbf{Churro (3B)}} & \rotatebox[origin=c]{90}{\textbf{\shortstack[c]{DeepSeek-\\OCR (3B)}}} & \rotatebox[origin=c]{90}{\textbf{\shortstack[c]{DeepSeek-\\OCR-2 (3B)}}} & \rotatebox[origin=c]{90}{\textbf{\shortstack[c]{LightOnOCR-2\\(1B)}}} & \rotatebox[origin=c]{90}{\textbf{\shortstack[c]{olmOCR-2\\(8B)}}} & \rotatebox[origin=c]{90}{\textbf{\shortstack[c]{QWEN3-VL\\(8B)}}} & \rotatebox[origin=c]{90}{\textbf{\shortstack[c]{RolmOCR\\(8B)}}} & \rotatebox[origin=c]{90}{\textbf{Azure Layout}} & \rotatebox[origin=c]{90}{\textbf{PeroOCR}} & \rotatebox[origin=c]{90}{\textbf{\shortstack[c]{Claude\\Opus 4.5}}} & \rotatebox[origin=c]{90}{\textbf{Gemini 3 Pro}} & \rotatebox[origin=c]{90}{\textbf{GPT-5.1}} & \rotatebox[origin=c]{90}{\textbf{\shortstack[c]{Mistral AI\\small}}} & \rotatebox[origin=c]{90}{\textbf{\shortstack[c]{Mistral AI\\medium}}} & \rotatebox[origin=c]{90}{\textbf{\shortstack[c]{Mistral AI\\large}}} & \rotatebox[origin=c]{90}{\textbf{Mistral OCR}} \\ \hline
        \textbf{ar} & 41.8 & 195.0 & 57.3 & 97.9 & 45.0 & 39.1 & 33.5 & 17.5 & 94.9 & \underline{16.9} & \textbf{13.8} & 62.2 & 154.5 & 63.5 & 69.1 & 19.4 \\
        \textbf{bg} & \underline{7.2} & 36.2 & 43.9 & 18.8 & 11.4 & 7.9 & 24.6 & 10.8 & 13.1 & 31.4 & \textbf{4.6} & 37.0 & 32.2 & 28.4 & 23.6 & 34.5 \\
        \textbf{bn} & 17.6 & 34.2 & 53.3 & 87.8 & 15.7 & 23.4 & 19.9 & 74.0 & 82.3 & \underline{8.8} & \textbf{6.5} & 20.8 & 25.0 & 21.2 & 20.8 & 23.0 \\
        \textbf{ca} & \textbf{9.4} & 216.9 & 67.4 & 247.9 & 15.3 & 19.0 & 18.2 & 32.1 & 47.0 & \underline{10.8} & 16.5 & 28.7 & 20.8 & 20.1 & 22.1 & 24.3 \\
        \textbf{cs} & 6.5 & 2.8 & 4.1 & 1.7 & 2.0 & 1.4 & 2.1 & 2.8 & \textbf{0.9} & 1.4 & \underline{1.1} & 6.6 & 2.8 & 3.1 & 2.8 & 1.6 \\
        \textbf{de} & \underline{51.4} & 177.8 & 119.4 & 196.9 & 106.5 & 78.1 & 106.5 & 70.4 & 63.0 & 53.6 & \textbf{33.9} & 66.7 & 73.6 & 81.3 & 69.8 & 67.8 \\
        \textbf{el} & \textbf{24.2} & 662.5 & 126.1 & 439.0 & 179.8 & 65.3 & 134.3 & 79.6 & 90.6 & 47.0 & \underline{45.8} & 72.9 & 524.4 & 78.3 & 91.6 & 84.2 \\
        \textbf{en} & 49.0 & 68.7 & 56.7 & 67.9 & 34.1 & \underline{29.3} & 58.2 & 32.4 & 68.7 & 57.4 & \textbf{17.3} & 64.1 & 219.0 & 54.2 & 53.7 & 57.9 \\
        \textbf{es} & 24.1 & 179.6 & 76.2 & 153.0 & 30.8 & 39.5 & 85.1 & 26.0 & 41.4 & \textbf{14.4} & \underline{15.3} & 39.3 & 44.8 & 48.2 & 42.2 & 27.3 \\
        \textbf{fa} & 206.5 & 321.9 & 106.53 & 192.8 & 180.7 & 56.2 & 47.8 & 55.1 & 97.7 & \underline{29.7} & \textbf{21.9} & 58.4 & 622.2 & 64.9 & 69.2 & 42.4 \\
        \textbf{fi} & 56.6 & 88.2 & 88.1 & 84.7 & 55.9 & 50.2 & 76.0 & \underline{32.2} & 52.9 & 75.8 & \textbf{19.4} & 80.9 & 87.7 & 82.4 & 77.0 & 75.2 \\
        \textbf{fr} & 16.9 & 97.8 & 32.9 & 123.4 & \underline{11.4} & 20.4 & 41.9 & 16.6 & 31.6 & 14.1 & \textbf{6.2} & 16.3 & 18.6 & 18.5 & 19.6 & 12.3 \\
        \textbf{he} & 239.4 & 178.9 & 89.2 & 197.5 & 115.1 & 89.2 & 128.0 & 106.9 & 93.7 & \underline{71.4} & \textbf{52.0} & 93.6 & 226.1 & 89.6 & 77.5 & 73.5 \\
        \textbf{hi} & \underline{8.2} & 24.3 & 62.8 & 198.2 & 17.1 & 15.2 & 17.5 & 32.4 & 94.4 & 14.3 & \textbf{7.8} & 32.3 & 19.2 & 20.8 & 48.6 & 33.7 \\
        \textbf{it} & 16.6 & 44.7 & 71.6 & 23.4 & 14.0 & 13.2 & 17.1 & 27.2 & 53.9 & \textbf{10.3} & \underline{10.4} & 39.7 & 21.5 & 20.3 & 24.0 & 16.2 \\
        \textbf{ja} & \underline{43.6} & 863.0 & 255.2 & 1011.9 & 403.6 & 64.2 & 197.1 & 51.9 & 102.8 & 62.3 & \textbf{35.2} & 90.9 & 100.4 & 126.0 & 223.7 & 62.4 \\
        \textbf{km} & \underline{84.5} & 559.8 & 94.5 & 1119.7 & 284.7 & 122.9 & 259.7 & 96.7 & 99.0 & 103.9 & \textbf{67.2} & 98.6 & 129.4 & 100.6 & 113.9 & 96.1 \\
        \textbf{la} & \underline{17.7} & 49.1 & 46.5 & 135.6 & 26.2 & 27.3 & 31.4 & 34.8 & 35.9 & 17.8 & \textbf{14.3} & 46.1 & 48.7 & 30.2 & 38.2 & 42.2 \\
        \textbf{nl} & \textbf{6.6} & 121.2 & 65.3 & 110.0 & 26.7 & 17.4 & 57.1 & 18.5 & 39.2 & 45.7 & \underline{13.7} & 45.1 & 55.7 & 51.3 & 51.2 & 49.0 \\
        \textbf{no} & 16.4 & 82.3 & 118.5 & 37.9 & 18.1 & 42.2 & 22.5 & 31.2 & 65.9 & \textbf{11.5} & \underline{12.3} & 30.8 & 39.1 & 37.6 & 61.7 & 17.7 \\
        \textbf{pl} & 28.7 & 89.8 & 89.7 & 59.9 & 39.1 & \underline{23.2} & 54.3 & 37.4 & 64.0 & 55.3 & \textbf{17.8} & 64.1 & 71.8 & 65.8 & 62.2 & 64.7 \\
        \textbf{pt} & 81.6 & 221.2 & 127.6 & 280.8 & 72.8 & 88.1 & 156.5 & 68.6 & 73.9 & \textbf{55.6} & \underline{59.1} & 84.4 & 1346.9 & 82.1 & 91.5 & 80.2 \\
        \textbf{ro} & \textbf{21.1} & 82.4 & 104.7 & 41.6 & 50.1 & 40.7 & 38.9 & 45.9 & 48.5 & \underline{32.1} & 41.3 & 67.9 & 46.4 & 39.7 & 54.2 & 40.5 \\
        \textbf{sa} & \textbf{39.1} & 238.7 & 224.7 & 188.0 & 190.5 & 50.8 & 72.3 & 46.6 & 95.3 & 43.6 & \underline{39.9} & 62.8 & 55.5 & 53.3 & 63.7 & 50.3 \\
        \textbf{sl} & \underline{1.7} & 5.0 & 6.5 & 4.3 & 4.6 & 2.9 & 4.4 & 5.1 & \textbf{1.4} & 3.6 & \underline{1.7} & 4.7 & 3.6 & 4.9 & 4.0 & 5.4 \\
        \textbf{sv} & 49.1 & 102.2 & 90.3 & 68.2 & 49.5 & \underline{29.2} & 60.1 & 44.8 & 71.6 & 52.4 & \textbf{18.2} & 56.5 & 68.5 & 63.0 & 61.9 & 62.0 \\
        \textbf{tr} & \textbf{29.7} & 270.1 & 77.3 & 198.3 & 585.9 & 112.4 & 172.1 & 60.4 & 96.6 & 59.3 & 63.6 & 69.5 & 98.0 & 106.3 & 97.8 & \underline{54.7} \\
        \textbf{vi} & \textbf{40.4} & 59.0 & 144.7 & 1666.2 & 49.9 & 52.7 & 46.3 & 57.6 & 100.8 & 77.0 & \underline{43.1} & 84.8 & 1098.8 & 119.7 & 142.3 & 86.4 \\
        \textbf{zh} & 177.7 & 138.4 & 164.9 & 297.1 & 104.9 & 83.8 & 97.3 & \underline{80.5} & 109.5 & 82.1 & \textbf{73.1} & 90.0 & 218.7 & 95.3 & 105.4 & 90.7 \\ \hline
        \textbf{Rank} & 3 & 16 & 14 & 15 & 7 & 4 & 8 & 5 & 11 & 2 & 1 & 9 & 13 & 10 & 12 & 6 \\ \hline
    \end{tabular}
\end{table}

%% file: tables/cer_handwritten_printed.tex
\begin{table}[ht!]
    \centering
    \fontsize{8pt}{8pt}\selectfont
    \caption{Character Error Rate (\%) for each model on printed and handwritten documents. Best values are highlighted in bold and second-best results are underlined.}
    \label{tab:results_printed_handwritten}
    \begin{tabular}{l|rr}
    \hline
        \textbf{Model} & \textbf{Printed} & \textbf{Handwritten} \\ \hline
        Churro (3B) & 41.0 & 58.1 \\ 
        DeepSeek-OCR (3B) & 72.3 & 234.5 \\
        DeepSeek-OCR-2 (3B) & 67.5 & 104.4 \\ 
        LightOnOCR-2 (1B) & 65.2 & 255.7 \\ 
        olmOCR-2 (8B) & 38.6 & 95.3 \\ 
        QWEN3-VL (8B) & 31.6 & 54.9 \\ 
        RolmOCR (8B) & 51.1 & 89.6 \\ \hline
        
        Azure Layout & \underline{30.8} & 49.1 \\
        PeroOCR & 53.2 & 73.9 \\ \hline

        Claude Opus 4.5 & 53.5 & \underline{31.2} \\ 
        Gemini 3 Pro & \textbf{17.6} & \textbf{28.5} \\ 
        GPT-5.1 & 58.9 & 59.5 \\
        Mistral AI small & 66.1 & 280.7 \\ 
        Mistral AI medium & 57.9 & 60.0 \\ 
        Mistral AI large & 56.6 & 64.3 \\ 
        Mistral OCR & 56.1 & 43.6 \\ \hline
    \end{tabular}
\end{table}

%% file: tables/dataset_computational_evaluation.tex
\begin{table}[ht!]
    \centering
    \fontsize{8pt}{8pt}\selectfont
    \caption{Documents selected for computational efficiency evaluation.}
    \label{tab:dataset_comp_eval}
    \begin{tabular}{l|lllrr}
    \hline
        \textbf{Dataset} & \textbf{Image type} & \textbf{Writing} & \textbf{Language} & \textbf{Characters} & \textbf{Words} \\ \hline
        Churro & Very long page & Handwritten & English & 20,695 & 3,830 \\ 
        Churro & Paragraph & Handwritten & Turkish & 737 & 137 \\ 
        Churro & Two columns & Handwritten (medieval) & Spanish & 2,504 & 499 \\
        IAM & Paragraph & Handwritten & English & 347 & 68 \\ 
        NorHand & Page & Handwritten & Norwegian & 2,169 & 396 \\ \hline
    \end{tabular}
\end{table}

%% file: tables/computational_evaluation.tex
\begin{table}[t!]
    \centering
    \fontsize{8pt}{8pt}\selectfont
    \caption{Inference time and memory consumption on the five selected documents. /100c denotes the average time for 100 characters, and /100w for 100 words. The last column indicates the number of times the model hallucinated during prediction. Best values are highlighted in bold and second-best results are underlined.}
    \label{tab:comp_eval}
    \begin{tabular}{l|rrrrr|rrr|r}
        \hline
        \multirow{2}{*}{\textbf{Model}} & \multicolumn{5}{c|}{\textbf{Time (seconds)}} & \multicolumn{3}{c|}{\textbf{Memory (GiB)}} & \multirow{2}{*}{\textbf{Hal.}} \\ 
        \textbf{} & \textbf{Min} & \textbf{Mean} & \textbf{Max} & \textbf{/100c} & \textbf{/100w} & \textbf{Min} & \textbf{Mean} & \textbf{Max} &  \\ 
        \hline
        Churro (3B) & 7.84 & 77.48 & 269.42 & 1.54 & 8.46 & 7.96 & 8.08& 8.12 & 0/5 \\ 
        Deepseek-OCR (3B) & 2.97 & 23.95 & 45.54 & 0.91 & 4.72 & 7.26 & 7.32 & 7.57 &  2/5 \\ 
        Deepseek-OCR-2 (3B) & 2.73 & 23.72 & 66.67 & 0.86 & 4.56 & 7.29 & 7.53 & 7.59 & 2/5 \\ 
        LightOnOCR-2 (1B) & 2.41 & 38.37 & 130.32 & 0.93 & 4.85 & 2.26 & 2.41 & 2.68 & 1/5 \\
        olmOCR-2 (8B) & 31.49 & 175.10 & 291.77 & 9.16 & 49.49 & 16.82 & 17.54 & 19.85 &0/5 \\ 
        QARI (2B) & 4.39 & 32.92 & 50.27 & 1.10 & 7.60 & 5.57 & 6.63 & 9.54 & 0/5 \\
        Qwen3-VL (8B) & 3.41 & 46.90 & 153.12 & 0.97 & 5.40 & 17.35 & 18.11 & 20.18 & 0/5 \\ 
        RolmOCR (8B) & 4.14 & 31.57 & 54.77 & 0.89 & 4.78 & 17.75 & 18.51& 20.62 & 0/5 \\ \hline

        Azure Layout & 5.21 & 7.98 & 18.30 & 0.14 & 0.76 & & & & 0/5 \\ 
        PeroOCR & \textbf{0.37} & \underline{2.51} & \underline{10.14} & \textbf{0.05} & \textbf{0.25} & 0.31 & 0.52& 0.79 & 0/5 \\ \hline

        Claude Opus 4.5 & 14.36 & 49.73 & 99.26 & 1.33 & 7.26 & & & & 0/5 \\ 
        Gemini 3 Pro & 43.13 & 105.16 & 143.78 & 1.84 & 10.31 & & & & 0/5 \\ 
        GPT-5.1 & 2.09 & 9.50 & 14.24 & 0.67 & 3.67 & & & & 1/5 \\ 
        Mistral AI small & 4.30 & 11.62 & 25.75 & 0.38 & 2.13 & & & & 1/5 \\ 
        Mistral AI medium & 3.58 & 15.02 & 31.05 & 0.64 & 3.50 & & & & 0/5 \\ 
        Mistral AI large & 5.35 & 30.62 & 60.50 & 0.92 & 5.30 & & & & 0/5 \\ 
        Mistral OCR & \underline{1.04} & \textbf{2.06} & \textbf{2.63} & \underline{0.12} & \underline{0.61} & & & & 0/5 \\ 
        \hline
    \end{tabular}
\end{table}